\documentclass{article}
\usepackage{spconf,amsmath,graphicx}

\usepackage{times}
\usepackage{epsfig}
\usepackage{graphicx}
\usepackage{amsmath}
\usepackage{amssymb}
\usepackage{booktabs}
\usepackage{comment}
\usepackage{subfigure}
\usepackage{caption}
\usepackage[normalem]{ulem}
\usepackage{multirow}

\usepackage{listings}
\usepackage{color}
 \usepackage[table,xcdraw]{xcolor}

\usepackage{mathtools}

\usepackage[pagebackref=false,breaklinks=true,letterpaper=true,colorlinks,bookmarks=false, linkcolor=blue, filecolor=blue, urlcolor=blue, citecolor=blue]{hyperref}

\title{Flexible-modal Deception Detection with Audio-Visual Adapter}
\name{\begin{tabular}{c}Zhaoxu Li\textsuperscript{1}, Zitong Yu\textsuperscript{1*\thanks{* Corresponding author}}, Nithish Muthuchamy Selvaraj\textsuperscript{1}, Xiaobao Guo\textsuperscript{1}, \\
Bingquan Shen\textsuperscript{2}, Adams Wai-Kin Kong\textsuperscript{1}, Alex Kot\textsuperscript{1}\end{tabular}}
\address{\textsuperscript{1}Nanyang Technological University, Singapore\qquad
\textsuperscript{2}DSO National Laboratories, Singapore}

\begin{document}
%
\maketitle
\begin{abstract}
Detecting deception by human behaviors is vital in many fields such as custom security and multimedia anti-fraud. Recently, audio-visual deception detection attracts more attention due to its better performance than using only a single modality. However, in real-world multi-modal settings, the integrity of data can be an issue (e.g., sometimes only partial modalities are available). The missing modality might lead to a decrease in performance, but the model still learns the features of the missed modality. In this paper, to further improve the performance and overcome the missing modality problem, we propose a novel Transformer-based framework with an Audio-Visual Adapter (AVA) to fuse temporal features across two modalities efficiently. Extensive experiments conducted on two benchmark datasets demonstrate that the proposed method can achieve superior performance compared with other multi-modal fusion methods under flexible-modal (multiple and missing modalities) settings. 
\end{abstract}

%
\begin{keywords}
Deception detection, multi-modal
\end{keywords}

 \vspace{-0.7em}
\section{Introduction}
 \vspace{-0.5em}

\label{sec:intro}

With the rapid development of the internet, security has become an important issue that can reduce millions of money being lost. An effective deception detection method can play a vital role and is widely used to protect people in the area of border security, anti-fraud, business negotiations and etc.

Early works in deception detection research were based on the psychological premises of deception \cite{depaulo2003cues,hirschberg2005distinguishing,levine2014theorizing,vrij2012eliciting,warren2009detecting}, such as physiological, visual, vocal, verbal, and behavioral cues. Deep learning pre-trained models have achieved comparable or even better performance than humans in many complex tasks so that deep convolutional neural networks (CNN)~\cite{he2016deep} and vision transformer (ViT)~\cite{dosovitskiy2020image} become mainstream in many computer vision tasks. In the deception detection area, existing models can be divided into single-modal and multi-modal models. For multi-modal deception detection, Gogate et al. \cite{gogate2017deep} presented a fusion model combining audio cues along with visual and textual cues for the first time. Karimi et al. \cite{karimi2018toward} proposed a robust method to capture rich information into a coherent end-to-end model. Wu et al. \cite{wu2018deception} studied the importance of different modalities like vision (low-level video features), audio (Mel-frequency Cepstral Coefficients), and text for this task. Mathur et al. \cite{mathur2020introducing} presented a novel method to analyze the discriminative power of facial affect for this task, and interpretable features from visual, vocal, and verbal modalities.

Most existing works~\cite{avola2019automatic,ding2019face,gogate2017deep} of multi-modal deception detection focus on the ideal situations in all modalities available at the deployment stage. However, in real-world practical scenarios, sometimes only partial modalities are available. In this case, the well-trained multi-modal models might perform even worse than the single-modal models. In this paper, we establish a flexible-modal deception detection benchmark and propose a novel fusion method for audio-visual deception detection. Our contributions include:
\begin{itemize}
\setlength\itemsep{-0.3em}
    \item We establish the first flexible-modal deception detection benchmark with both intra- and cross-dataset testings under both full multi-modal scenario (i.e., \textit{Vision+Audio}), and two missing-modal scenarios (\textit{Missing Vision}, \textit{Missing Audio}).
    \item We propose a novel Transformer-based framework with an Audio-Visual Adapter (AVA) module to fuse features across modalities efficiently. AVA significantly improves the performance of Transformer baselines in both flexible-modal intra- and cross-testings.
\end{itemize}
\begin{figure*}[t]
\centering
\includegraphics[scale=0.58]{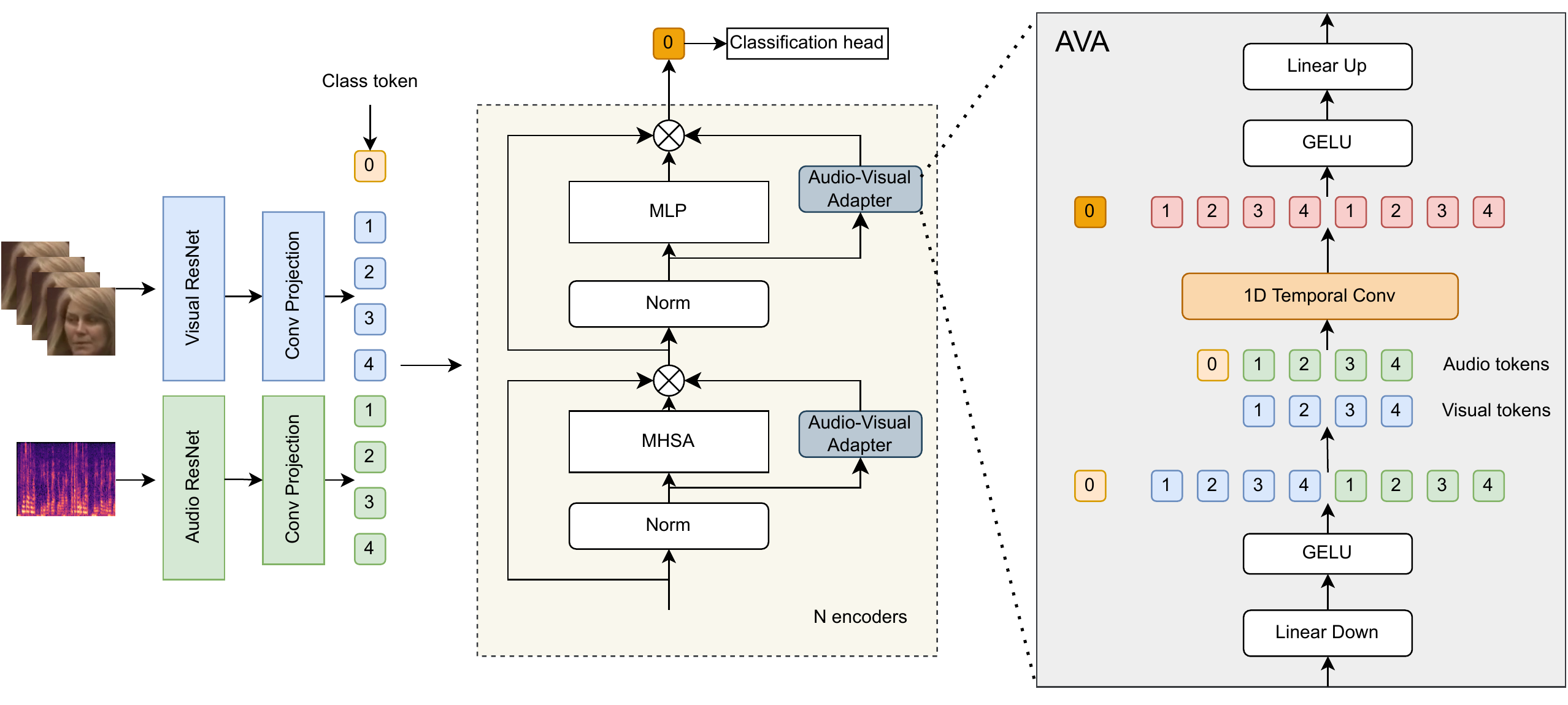}
\vspace{-1.3em}
  \caption{\small{
  The framework of the multi-modal deception detection with Audio-Visual Adapter (AVA). `MHSA’ and `MLP’ are short for the multi-head self-attention and multi-layer perception, respectively.}
  }
 \vspace{-1.0em}
\label{fig:network}

\end{figure*}
 \vspace{-1.4em}
\section{Methodology}
 \vspace{-0.5em}
\label{sec:method}

To take advantage of the powerful ImageNet pre-trained ViT and ResNet models, we propose a novel multi-modal deception detection framework to efficiently fine-tune the pre-trained models with audio-visual temporal adapters while fixing all pre-trained ViT weights. In this section, we will first introduce the whole model structure in Sec. \ref{sec2.1}. It consists of two unshared ResNet18 for visual and audio feature extraction, and then both visual and audio features are projected into temporal tokens and passed forward transformer encoders with global audio-visual attention. Besides, we will introduce the Audio-Visual Adapters in Sec. \ref{sec2.2}, which can plug in transformer encoders and efficiently aggregate the audio-visual cues among their local temporal neighborhoods.

\vspace{-0.8em}
\subsection{Multimoal Deception Detection Framework}
\vspace{-0.6em}

\label{sec2.1}
As illustrated in the left part of Fig. \ref{fig:network}, the $X_{A}$ and $X^{i}_{V}$ (i = 1,..., $T$) respectively denote audio and visual inputs, where $T$ is the frame numbers of visual input. After the feature extractor, the outputs can be formulated as
\vspace{-0.6em}
\begin{equation} 
\vspace{-0.6em}
\begin{split}
&F^{i}_{V} = \mathrm{RN_{V}(X^{i}_{V})},\\
&F_{A} = \mathrm{RN_{A}(X_{A})},
\vspace{-0.1em}
\end{split}
\end{equation}
where $RN_{V}$ and $RN_{A}$ are ImageNet pre-trained Visual ResNet model and Audio ResNet model of Fig.\ref{fig:network}. The $T$-frame visual features $F^{i}_{V}$ are concatenated in the time domain to form $F_{V}$. Then two Conv Projections $C^{A}_{proj}$ and $C^{V}_{proj}$ via convolutional projection with different sizes of the temporal kernel to align the visual and audio modalities along the time dimension. Specifically, the concatenated visual features $F_{V}$ and audio features $F_{A}$ are passed over $C^{V}_{proj}$ and $C^{A}_{proj}$ to generate the visual tokens $T_{V}$ and audio tokens $T_{A}$, respectively. All visual and audio tokens and a learnable class token $T_{C}$ are concatenated to extract the cross-modal feature and added with position embeddings. 
Then $N$ transformer blocks $E^{j}_{trans}$(j = 1, ..., $N$) are cascaded for global audio-visual feature interaction. Each transformer block is formed by two Layer Norm (LN) layers, a Multi-Head Self-Attention (MHSA) module, Audio-Visual Adapters (AVA), and a Multi-Layer Perception (MLP) module. The output $[T^{F}_{C},T^{F}_{V},T^{F}_{A}]$ of the transformer can be formalized as follows
\vspace{-0.6em}
\begin{equation} 
\vspace{-0.6em}
\begin{split}
&T_{V} = \mathrm{C^{V}_{proj}(Concat(F^{i}_{V}))}, (i = 1, ..., T ),\\
&T_{A} = \mathrm{C^{A}_{proj}(F_{A})},\\
&[T^{F}_{C},T^{F}_{V},T^{F}_{A}] = \mathrm{E_{trans}(Concat(T_{C},T_{V},T_{A}))}.
\vspace{-0.1em}
\end{split}
\end{equation}
Finally, $T^{F}_{C}$ is sent to a classification head $E_{head}$ for binary truth/lie prediction.

\vspace{-0.6em}
\subsection{Audio-Visual Adapter}
\vspace{-0.6em}

\label{sec2.2}
To alleviate the overfitting issues of the transformer finetuning, we propose to train the Audio-Visual Adapter (AVA) to fuse the visual and audio tokens efficiently while fixing all the pre-trained parameters of transformer encoders $E_{trans}$. These adapters are placed in parallel with MHSA and MLP modules which are shown in the right part of Fig.\ref{fig:network}. Unlike channel-wise NLP adapter \cite{houlsby2019parameter} and spatial-wise Conv-adapter \cite{jie2022convolutional}, our method is designed to capture the temporal multi-modal feature. The concatenated audio-visual tokens are first reshaped to align the visual and audio tokens which are at the same time. Then a 1D convolution layer with temporal kernel size $k$ and padding $k//2$ is used to aggregate the local temporal multi-modal features. The output $A(T)$ of AVA can be formulated as
\vspace{-0.6em}
\begin{equation} 
\vspace{-0.6em}
\begin{split}
&T^{Re}_{ALL} = \mathrm{Reshape(GELU(LD(Concat(T_{C},T_{V},T_{A}))))},\\
&T_{F} = \mathrm{Conv(T^{Re}_{ALL})},\\
&A(T_{F}) = \mathrm{LU(GELU(Reshape(T_{F})))},
\vspace{-0.1em}
\end{split}
\end{equation}
where $LD$, $LU$, and $Conv$ are linear down, linear up, and 1D temporal Conv layers in Fig.\ref{fig:network}, and $T^{Re}_{ALL}$ and $T_{F}$ represent the reshaped temporally-aligned tokens and fused tokens, respectively.
Specifically, the input tokens $T^{j}$ and the output $E^{j}_{T}$ of each encoders with AVA can be formulated as:
\vspace{-0.6em}
\begin{equation} 
\vspace{-0.6em}
\begin{split}
&T^{j}_{'} = \mathrm{T^{j}+MHSA(LN(T^{i}))+A(T^{j})},\\
&E^{j}_{T} = \mathrm{T^{j}_{'}+LN(MLP(T^{j}_{'}))+A(T^{j}_{'})}.
\vspace{-0.1em}
\end{split}
\end{equation}


\begin{table}[t]

\centering
\caption{\small{Results of intra-dataset testings on RLtrial. `A' and `V' are short for audio and vision modality, respectively.}}
\vspace{-0.8em}
\label{tab:my-rl}
\resizebox{\columnwidth}{!}{%
\large
\begin{tabular}{|cc|cc|ccc|}
\hline
\multicolumn{2}{|c|}{\multirow{2}{*}{Method}} &
  \multicolumn{1}{c|}{\multirow{2}{*}{Train}} &
  \multirow{2}{*}{Test} &
  \multicolumn{3}{c|}{RLtrial} \\ \cline{5-7} 
\multicolumn{2}{|c|}{} &
  \multicolumn{1}{c|}{} &
   &
  \multicolumn{1}{c|}{ACC(\%)} &
  \multicolumn{1}{c|}{AUC(\%)} &
  F1 (\%) \\ \hline
\multicolumn{1}{|c|}{\multirow{4}{*}{Single}} & \multirow{2}{*}{ResNet}                            & A    & A    & 64.89          & 65.14          & 72.30 \\
\multicolumn{1}{|c|}{}                        &                                                    & V    & V    & 59.48          & 59.34          & 68.97          \\ \cline{2-7} 
\multicolumn{1}{|c|}{}                        & \multirow{2}{*}{ResNet+Transformer}                & A    & A    & 65.39          & 65.27          & 71.45          \\
\multicolumn{1}{|c|}{}                        &                                                    & V    & V    & 61.32          & 61.03          & 69.54          \\ \hline
\multicolumn{1}{|c|}{\multirow{15}{*}{Unified}} &
  \multirow{3}{*}{ResNet+Transformer+Concat} &
  V\&A &
  V\&A &
  68.99 &
  73.19 &
  68.92 \\
\multicolumn{1}{|c|}{}                        &                                                    & V\&A & A    & 61.87          & 66.87          & 61.91          \\
\multicolumn{1}{|c|}{}                        &                                                    & V\&A & V    & 66.78          & 62.88          & 66.53          \\ \cline{2-7} 
\multicolumn{1}{|c|}{}                        & \multirow{3}{*}{ResNet+Transformer+SE-Concat} & V\&A & V\&A & 67.75          & 71.10          & 67.68          \\
\multicolumn{1}{|c|}{}                        &                                                    & V\&A & A    & 63.30          & 70.30          & 63.07          \\
\multicolumn{1}{|c|}{}                        &                                                    & V\&A & V    & 66.91          & 72.55          & 66.82          \\ \cline{2-7} 
\multicolumn{1}{|c|}{} &
  \multirow{3}{*}{ResNet+Transformer+CMFL} &
  V\&A &
  V\&A &
  64.89 &
  64.94 &
  70.63 \\
\multicolumn{1}{|c|}{}                        &                                                    & V\&A & A    & 61.93          & 61.76          & 70.29          \\
\multicolumn{1}{|c|}{}                        &                                                    & V\&A & V    & 64.62          & 64.45          & 68.73          \\ \cline{2-7} 
\multicolumn{1}{|c|}{}                        & \multirow{3}{*}{ResNet+Transformer+Prompt}         & V\&A & V\&A & 66.37          & 71.26          & 66.44          \\
\multicolumn{1}{|c|}{}                        &                                                    & V\&A & A    & 66.94          & 72.97          & 67.03          \\
\multicolumn{1}{|c|}{}                        &                                                    & V\&A & V    & 67.46          & 73.43          & 67.41          \\ \cline{2-7} 
\multicolumn{1}{|c|}{} &
  \multirow{3}{*}{\textbf{Ours (ResNet+Transformer+AVA)}} &
  V\&A &
  V\&A &
  \textbf{74.62} &
  \textbf{78.30} &
  \textbf{74.52} \\
\multicolumn{1}{|c|}{}                        &                                                    & V\&A & A    & \textbf{71.32} & \textbf{74.39} & \textbf{71.26 }         \\
\multicolumn{1}{|c|}{}                        &                                                    & V\&A & V    & \textbf{71.93} & \textbf{76.31} & \textbf{71.92} \\ \hline
\end{tabular}%
}
\vspace{-1.6em}

\end{table}

\vspace{-1.2em}
\section{Experiment}
\label{sec:experiment}
 \vspace{-0.2em}

\vspace{-0.6em}
\subsection{Flexible-Modal Benchmark}
 \vspace{-0.4em}
\noindent\textbf{Datasets.}\quad Two multi-modal deception detection datasets are used in the flexible-modal benchmark. \textbf{RLtrial}: Real-life trial~\cite{perez2015deception} consists of 121 deceptive and truthful video clips with 27.7 seconds for deceptive and 28.3 seconds for truthful clips on average respectively, from real court trials. The data includes approximately 21 female and 35 unique male speakers ranging between 16 and 60 years.
\textbf{BOL}: Bag of lies~\cite{gupta2019bag} consists of 325 annotated videos (163 truth and 162 lie) recorded by 35 subjects. It includes 4 types of data such as video, audio, EEG, and Eye gaze. In our experiments, we focus on audio-visual deception detection.

 \vspace{0.2em}
\noindent\textbf{Protocols and evaluation metrics.}\quad We evaluate models under flexible-modal protocols on both intra-dataset (5-fold for RLtrial and 3-fold for BOL) and cross-dataset (trained on RLtrial/BOL tested on BOL/RLtrial) testings. In terms of flexible-modal settings, we train the multi-modal models with both audio and video modalities, and evaluate them under \textit{Vision}-based (V), \textit{Audio}-based (A) single-modal, and \textit{Vision+Audio}-based (V\&A) scenarios. Compared with training unified multi-modal models (see `Unified' in Tables~\ref{tab:my-rl},~\ref{tab:my-Bol} and~\ref{tab:crosstable}) with audio-visual modalities, we also consider single-modal results (see `Single' in Tables~\ref{tab:my-rl},~\ref{tab:my-Bol} and~\ref{tab:crosstable}) that train separate models with only audio or video data. We evaluate our method and other fusion methods with Accuracy (ACC), F1 score (F1), and Area Under Curve (AUC).

\vspace{-1.2em}
\subsection{Implementation Details}
 \vspace{-0.4em}
 
All the experiments are implemented with Pytorch on one NVIDIA RTX A5000 GPU. The batch size is 8, and we use the SGD optimizer with the learning rate (lr) of 1e-4, the weight delay of 5e-5, and the momentum of 0.9. We use StepLR as the learning rate scheduler with a step size of 20 and the gamma of 0.1. ResNet18 and the encoder of ViT are fine-tuned based on the ImageNet/ImageNet-21K pre-trained models with 25 epochs of training. The missing modalities experiments are simply blocked inputs as zeros in the testing phase of Protocols 1 and 2.

\begin{table}[t]
\centering
\caption{ \small{Results of intra-dataset testings on BOL. }}
\vspace{-0.8em}
\label{tab:my-Bol}
\resizebox{\columnwidth}{!}{%
\large
\begin{tabular}{|cc|cc|ccc|}
\hline
\multicolumn{2}{|c|}{\multirow{2}{*}{Method}} & \multicolumn{1}{c|}{\multirow{2}{*}{Train}} & \multirow{2}{*}{Test} & \multicolumn{3}{c|}{BOL} \\ \cline{5-7} 
\multicolumn{2}{|c|}{} &
  \multicolumn{1}{c|}{} &
   &
  \multicolumn{1}{c|}{ACC(\%)} &
  \multicolumn{1}{c|}{AUC(\%)} &
  F1(\%) \\ \hline
\multicolumn{1}{|c|}{\multirow{4}{*}{Single}} &
  \multirow{2}{*}{ResNet} &
  A &
  A &
  54.79 &
  55.87 &
  64.27 \\
\multicolumn{1}{|c|}{} &
   &
  V &
  V &
  54.07 &
  54.42 &
  64.66 \\ \cline{2-7} 
\multicolumn{1}{|c|}{} &
  \multirow{2}{*}{ResNet+Transformer} &
  A &
  A &
  57.44 &
  58.50 &
  67.11 \\
\multicolumn{1}{|c|}{} &
   &
  V &
  V &
  54.35 &
  54.67 &
  66.29 \\ \hline
\multicolumn{1}{|c|}{\multirow{15}{*}{Unified}} &
  \multirow{3}{*}{ResNet+Transformer+Concat} &
  V\&A &
  V\&A &
  57.57 &
  57.69 &
  67.59 \\
\multicolumn{1}{|c|}{} &
   &
  V\&A &
  A &
  55.65 &
  57.68 &
  61.81 \\
\multicolumn{1}{|c|}{} &
   &
  V\&A &
  V &
  55.95 &
  54.59 &
  38.63 \\ \cline{2-7} 
\multicolumn{1}{|c|}{} &
  \multirow{3}{*}{ResNet+Transformer+SE-Concat} &
  V\&A &
  V\&A &
  59.19 &
  59.06 &
  67.21 \\
\multicolumn{1}{|c|}{} &
   &
  V\&A &
  A &
  55.30 &
  56.40 &
  56.94 \\
\multicolumn{1}{|c|}{} &
   &
  V\&A &
  V &
  55.84 &
  \textbf{56.84} &
  44.03 \\ \cline{2-7} 
\multicolumn{1}{|c|}{} &
  \multirow{3}{*}{ResNet+Transformer+CMFL} &
  V\&A &
  V\&A &
  53.50 &
  55.30 &
  65.68 \\
\multicolumn{1}{|c|}{} &
   &
  V\&A &
  A &
  52.60 &
  54.64 &
  65.35 \\
\multicolumn{1}{|c|}{} &
   &
  V\&A &
  V &
  53.16 &
  52.27 &
  43.32 \\ \cline{2-7} 
\multicolumn{1}{|c|}{} &
  \multirow{3}{*}{ResNet+Transformer+Prompt} &
  V\&A &
  V\&A &
  59.54 &
  59.98 &
  \textbf{68.91} \\
\multicolumn{1}{|c|}{} &
   &
  V\&A &
  A &
  56.73 &
  54.74 &
  66.84 \\
\multicolumn{1}{|c|}{} &
   &
  V\&A &
  V &
  54.51 &
  53.46 &
  \textbf{67.85} \\ \cline{2-7} 
\multicolumn{1}{|c|}{} &
  \multirow{3}{*}{\textbf{Ours (ResNet+Transformer+AVA)}} &
  V\&A &
  V\&A &
  \textbf{61.20} &
  \textbf{61.48} &
  68.30 \\
\multicolumn{1}{|c|}{} &
   &
  V\&A &
  A &
  \textbf{57.96} &
  \textbf{57.98} &
  \textbf{68.09} \\
\multicolumn{1}{|c|}{} &
   &
  V\&A &
  V &
  \textbf{56.56} &
  54.48 &
  66.93 \\ \hline
\end{tabular}%
}
\vspace{-1.6em}
\end{table}

\begin{table*}[t]
\centering
 \vspace{-0.2em}
\caption{ \small{Results of cross-dataset testings between RLtrial and BOL.}}
 \vspace{-0.8em}
\label{tab:crosstable}
\resizebox{0.67\textwidth}{!}{
\large
\begin{tabular}{|lc|c|c|ccc|ccc|}
\hline
\multicolumn{2}{|c|}{\multirow{2}{*}{Method}} & \multirow{2}{*}{Train} & \multirow{2}{*}{Test} & \multicolumn{3}{c|}{Train on RL Test on BOL} & \multicolumn{3}{c|}{Train on BOL Test on RL} \\ \cline{5-10} 
\multicolumn{2}{|c|}{} &  &  & \multicolumn{1}{c|}{ACC(\%)} & \multicolumn{1}{c|}{AUC(\&)} & F1(\%) & \multicolumn{1}{c|}{ACC(\%)} & \multicolumn{1}{c|}{AUC(\&)} & F1(\%) \\ \hline
\multicolumn{1}{|l|}{\multirow{4}{*}{Single}} & \multirow{2}{*}{ResNet} & A & A & 50.46 & 50.50 & 43.11 & 47.66 & 47.22 & 28.92 \\
\multicolumn{1}{|l|}{} &  & V & V & 51.40 & 50.96 & 26.51 & 50.15 & 50.02 & 65.09 \\ \cline{2-10} 
\multicolumn{1}{|l|}{} & \multirow{2}{*}{ResNet+Transformer} & A & A & 52.62 & 52.51 & 64.35 & 51.40 & 51.15 & 52.99 \\
\multicolumn{1}{|l|}{} &  & V & V & 52.92 & 52.85 & 67.36 & 54.21 & 53.79 & 62.82 \\ \hline
\multicolumn{1}{|l|}{\multirow{15}{*}{Unified}} & \multirow{3}{*}{ResNet+Transformer+Concat} & V\&A & V\&A & 50.76 & 50.63 & 66.80 & 53.27 & 53.05 & 54.87 \\
\multicolumn{1}{|l|}{} &  & V\&A & A & 50.15 & 50.30 & 62.86 & 56.07 & 56.15 & 59.13 \\
\multicolumn{1}{|l|}{} &  & V\&A & V & 54.77 & 54.78 & 66.94 & 54.21 & 53.79 & 25.71 \\ \cline{2-10} 
\multicolumn{1}{|l|}{} & \multirow{3}{*}{ResNet+Transformer+SE-Concat} & V\&A & V\&A & 50.15 & 50.26 & 41.77 & 50.46 & 50.30 & 42.42 \\
\multicolumn{1}{|l|}{} &  & V\&A & A & 50.15 & 50.31 & 12.5 & 51.40 & 51.22 & 39.53 \\
\multicolumn{1}{|l|}{} &  & V\&A & V & 51.69 & 51.66 & 66.26 & 58.88 & 58.49 & 66.25 \\ \cline{2-10} 
\multicolumn{1}{|l|}{} & \multirow{3}{*}{ResNet+Transformer+CMFL} & V\&A & V\&A & 50.46 & 50.58 & 66.80 & 52.34 & 51.90 & \textbf{64.56} \\
\multicolumn{1}{|l|}{} &  & V\&A & A & 51.69 & 51.59 &\textbf{ 66.94 }& 51.40 & 50.94 & \textbf{65.41} \\
\multicolumn{1}{|l|}{} &  & V\&A & V & 51.38 & 51.53 & 66.80 & 53.27 & 52.87 & 64.56 \\ \cline{2-10} 
\multicolumn{1}{|l|}{} & \multirow{3}{*}{ResNet+Transformer+Prompt} & V\&A & V\&A & 50.15 & 50.00 &\textbf{ 66.80} & 52.34 & 52.11 & 37.04 \\
\multicolumn{1}{|l|}{} &  & V\&A & A & 50.15 & 50.00 & 66.80 & 53.27 & 53.07 & 40.48 \\
\multicolumn{1}{|l|}{} &  & V\&A & V & 52.31 & 52.24 & 66.80 & 54.21 & 54.37 & \textbf{67.52} \\ \cline{2-10} 
\multicolumn{1}{|l|}{} & \multirow{3}{*}{\textbf{Ours (ResNet+Transformer+AVA)}} & V\&A & V\&A & \textbf{58.77} & \textbf{58.73} & 63.19 & \textbf{59.81} & \textbf{59.59} & 53.03 \\
\multicolumn{1}{|l|}{} &  & V\&A & A & \textbf{55.08} & \textbf{55.07} & 63.91 & \textbf{58.87} & \textbf{58.68} & 58.91 \\
\multicolumn{1}{|l|}{} &  & V\&A & V & \textbf{55.38} & \textbf{55.29} & \textbf{67.51} & \textbf{60.75} & \textbf{60.67} & 66.25 \\ \hline
\end{tabular}}
\vspace{-1.2em}
\end{table*}

 \vspace{0.2em}
\noindent\textbf{Data pre-processing.}\quad For each video clip, we uniformly sampled $T$ = 20 images then use MTCNN face detector \cite{zhang2016joint} to crop the face areas. After the cropping process, all images are normalized and resized to 224 × 224. For audio data, we transfer the raw audio to the Mel spectrogram image with the size of 640 × 480 using torchaudio. 

 \vspace{0.2em}
\noindent\textbf{Model details.}
We use pre-trained ResNet18~\cite{he2016deep} as the backbone for the single-modal feature extraction. Specifically, when cascaded with transformer blocks, we tokenize the $T$-frame face features from temporal dimensions 512 to 768, and also tokenize the audio spectrogram features (512 × $T$ × 15 for dimension × time × frequency) into size 512 × $T$ using global convolution on frequency dimension. We then utilize the first 8 transformer encoder layers ($N$=8) from ImageNet-21K pre-trained ViT model for global temporal feature refinement. For the AVA, temporal kernel size $k$=5 is utilized. 

 \vspace{0.2em}
\noindent\textbf{Fusion baselines.}  Besides fusion with AVA, we also compare with other fusion baselines: 1) \textbf{Concat}: Concatenate two features in the channel domain and then aggregate the multi-modal heterogeneous features with a lightweight fusion operator.
2) \textbf{SE-Concat}: Squeeze-and-excitation (SE) module~\cite{hu2018squeeze} is utilized in each independent modality branch first. With the channel-wise self-calibration via the SE module, the refined features are then concatenated. 
3) \textbf{CMFS}: Cross-modal focal loss \cite{george2021cross} is used to modulate the loss contribution of each channel as a function of the confidence of individual channels. 
4) \textbf{Prompt}: Visual Prompt Tuning method \cite{jia2022visual} introduces a small amount of task-specific learnable tokens while freezing the entire pre-trained transformer blocks during deception detection training.

\begin{figure*}[t]
\centering
\includegraphics[scale=0.22]{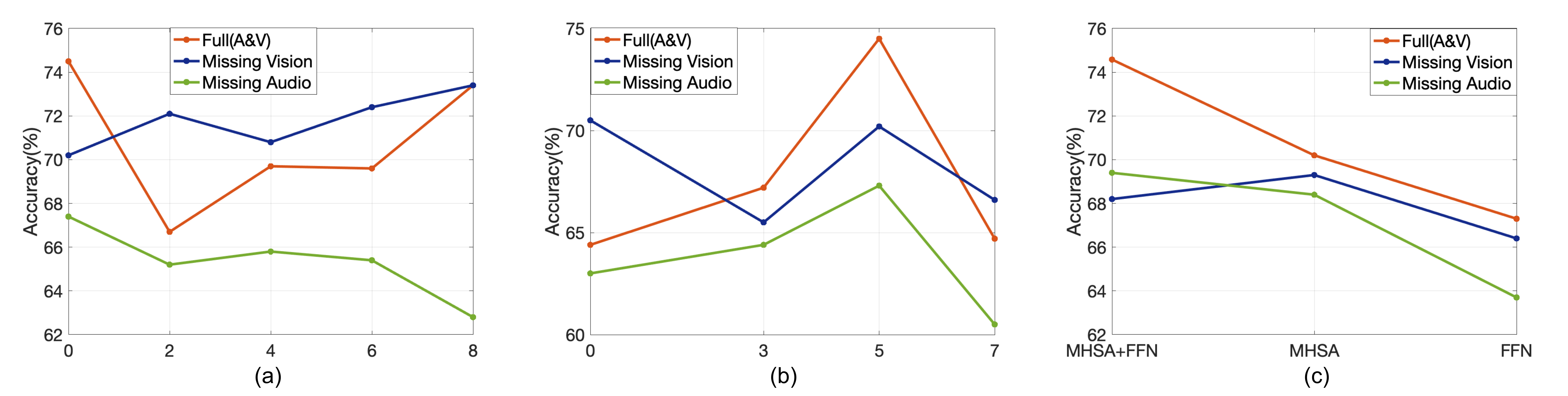}
\vspace{-1.2em}
  \caption{\small{
  Ablation of the (a) number of encoders without AVA; (b) temporal kernel size of AVA; (c) position of AVA.}
  }
 \vspace{-1.2em}
\label{fig:ablation}
\end{figure*}

\vspace{-1.2em}
\subsection{Intra Testing}
\vspace{-0.5em}
The experimental results of flexible-modal intra-dataset testing on RLtrial and BOL datasets are shown in Table~\ref{tab:my-rl} and Table~\ref{tab:my-Bol}, respectively.
It is clear that 1) models with audio modality usually perform better than Visual modality; and 2) the separate trained `ResNet+Transformer (A or V)' achieve worse performance than multi-modal models with two modalities (A\&V), indicating the necessity of multi-modal cues for deception detection.

 \vspace{0.2em}
 \noindent\textbf{Impact of fusion modules.}\quad As we can see from the `Unified' block in Table~\ref{tab:my-rl} and Table~\ref{tab:my-Bol}, compared with the single-modal models, most of the fusion methods can increase the performance.  In contrast, we can find from the results `ResNet+Transformer+AVA' that the proposed AVA improves the accuracy remarkably (with  61.20\% and 74.62\% for testing on BOL and testing on RL, respectively).

 \vspace{0.2em}
\noindent\textbf{Results of missing modalities.}\quad The missing-modal testings indicate that the multi-modal model can benefit from more modalities, but it might overfit on a better modal when missing modality. The results of two datasets tests of our method are not better than the fusion result when one modality is missed but still better than the single-modal model, indicating the model can actually learn the complementary features across modalities and also alleviates the overfitting problem.

 \vspace{-1.2em}
\subsection{Cross Testing}
 \vspace{-0.6em}
Table~\ref{tab:crosstable} shows the results of flexible-modal cross-dataset testings between RLtrial and BOL. Due to the domain shifts (e.g., environments and stakes) between two datasets, the performance of both separate and unified models are unsatisfactory.

 \vspace{0.2em}
\noindent\textbf{Impact of fusion modules.}\quad The results shows that other fusion methods cannot bring obvious benefits than the single modal model on cross-dataset testings. Compared with other fusion methods, the proposed AVA is suitable with multi-modal architectures both on two cross-dataset tests and improves the testing accuracy results significantly.

 \vspace{0.2em}
\noindent\textbf{Impact of missing modalities.}\quad Similar to intra-dateset testings, the fusion methods cannot achieve a satisfactory accuracy under missing modality scenarios due to the lack of partial modality cues and overfitting on partial/full modalities. In contrast, the proposed method can still have a reasonable accuracy under two missing-modal scenarios.

 \vspace{-1.2em}
\subsection{Ablation Study}
\vspace{-0.5em}
We also conduct elaborate ablation studies of AVA on the intra-dataset testings of RLtrail.

 \vspace{0.2em}
\noindent\textbf{Number of encoders without AVA.}\quad As shown in Fig.3(a), our model performs the best when all eight encoders assemble with AVA. The performance drops when more encoder layers are without AVA. Fewer encoder layers without AVA might lead to modality-aware overfitting (e.g., poor performance when missing audio modality), although the fusion performance is getting better.

 \vspace{0.2em}
\noindent\textbf{Kernel size of AVA.}\quad The effect of 1D convolutional kernel size $k$ in AVA is presented in Fig.3(b). AVA with $k$=5 achieves the best performance on both full and missing modalities settings. The performance dropped when using larger or smaller kernel sizes due to the overfitting or limited local temporal contexts, respectively. Note that AVA outperforms the NLP adapter \cite{houlsby2019parameter} (i.e., $k$=0) by a large margin. 

 \vspace{0.2em}
 
\noindent\textbf{Position of AVA.}\quad The results of AVA positions inside the transformer encoders are illustrated in Fig.3(c). It is clear that the model works best when assembling AVA parallelly on both MHSA and FFN layers. The fusion performance of MHSA was better than those placed in FFN layers. For missing modalities, the performance of MHSA+FFN is close to MHSA but obviously better than those placed in FFN.

 \vspace{-0.8em}
\section{Conclusion} 
\vspace{-0.6em}

In this paper, we propose a novel Transformer-based framework with Audio-Visual Adapter modules for multi-modal deception detection. We also provide sufficient baselines using other deep learning methods and feature fusion strategies for flexible-modal deception detection. In the future, we will investigate the scenarios when only partial modality data are available at the training stage.

\vspace{0.2em}
\noindent\textbf{Acknowledgment}  \quad This work was carried out at the Rapid-Rich Object Search (ROSE) Lab, Nanyang Technological University, Singapore. The research is supported by the DSO National Laboratories, under the project agreement No. DSOCL21238.



\clearpage

\bibliographystyle{IEEEbib}
\bibliography{refs}

\end{document}